%% file: acl2021.tex
\documentclass[11pt,a4paper]{article}
\usepackage[hyperref]{acl2021}
\usepackage{times}
\usepackage{latexsym}

\usepackage{microtype}

\aclfinalcopy 

\usepackage{paralist}
\usepackage{multirow}
\usepackage{booktabs}

\newcommand{\method}{NeurST\space}

\newcommand{\tabincell}[2]{\begin{tabular}{@{}#1@{}}#2\end{tabular}}

\title{NeurST: Neural Speech Translation Toolkit}

\author{Chengqi Zhao \quad Mingxuan Wang \quad Qianqian Dong \quad Rong Ye \quad Lei Li \\
  ByteDance AI Lab, Shanghai, China \\
  \small \texttt{\{zhaochengqi.d,wangmingxuan.89,dongqianqian,yerong,lileilab\}@bytedance.com} }

\begin{document}
\maketitle

\input{00-abstract}

\section{Introduction}
\label{sec:intro}
\input{01-introduction}

\input{02-design}

\input{03-experiment}

\input{04-conclusion}

\bibliography{anthology,custom}
\bibliographystyle{acl_natbib}


\end{document}

%% file: 00-abstract.tex
\begin{abstract}
\method is an open-source toolkit for neural speech translation. 
The toolkit mainly focuses on end-to-end speech translation, which is easy to use, modify, and extend to advanced speech translation research and products. 
\method aims at facilitating the speech translation research for NLP researchers and building reliable benchmarks for this field. It provides step-by-step recipes for feature extraction, data preprocessing, distributed training, and evaluation. In this paper, we will introduce the framework design of \method and show experimental results for different benchmark datasets, which can be regarded as reliable baselines for future research.
The toolkit is publicly available at \url{https://github.com/bytedance/neurst} and we will continuously update the performance of \method with other counterparts and studies at \url{https://st-benchmark.github.io/}.
\end{abstract}

%% file: 01-introduction.tex
Speech translation~(ST), which translates audio signals of speech in one language into text in a foreign language, is a hot research subject nowadays and has widespread applications, like cross-language videoconferencing or customer support chats. 

Traditionally, researchers build a speech translation system via a cascading manner, including an automatic speech recognition~(ASR) and a machine translation~(MT) subsystem~\cite{ney1999,casacuberta2008,kumar2014}. Cascade systems, however, suffer from error propagation problems, where an inaccurate ASR output would theoretically cause translation errors. 
Owing to recent progress of sequence-to-sequence modeling for both neural machine translation~(NMT)~\cite{bahdanau2015,luong2015,vaswani2017} and end-to-end speech recognition~\cite{chan2016,chiu2018,dong2018}, it becomes feasible and efficient to train an end-to-end direct ST model~\cite{berard2016,Duong2016,weiss2017}. This end-to-end fashion attracts much attention due to its appealing properties: \textit{a}) modeling without intermediate ASR transcriptions obviously alleviates the propagation of errors; \textit{b}) a single and unified ST model is beneficial to deployment with lower latency in contrast to cascade systems. 

Recent studies show that end-to-end ST models achieve promising performance and are comparable with cascaded models~\cite{ansari2020}. The end-to-end solution has great potential to be the dominant technology for speech translation, however challenges remain. 
The first is about benchmarks. Many ST studies conduct experiments on different datasets. ~\citet{liu2019end} evaluate the method on TED English-Chinese; and ~\citet{dong2020ted} use \textit{libri-trans} English-French and  IWSLT2018 English-German dataset; and ~\citet{wu2020self} show the results on CoVoST dataset and the FR/RO portions of MuST-C dataset. Different datasets make it difficult to compare the performance of their approaches. 
Further, even for the same dataset, the baseline results are not necessarily kept consistent. Take the \textit{libri-trans} English-French dataset as an example.  ~\citet{dong2020ted} report the pre-trained baseline as 15.3 and the result of ~\citet{liu2019end} is 14.3 in terms of tokenized BLEU, while~\citet{inaguma2020} report 15.5 (detokenized BLEU). 
The mismatching baseline results in an unfair comparison on the improvements of their approaches.
We think one of the primary reasons is that the preprocessing of audio data is complex, and the ST model training involves many tricks, such as pre-training and data augmentation.  

Therefore a reproducible and reliable benchmark is required. In this work, we present \method, a toolkit for easily building and training end-to-end ST models, as well as end-to-end ASR and NMT for cascade systems. 
We implement state-of-the-art Transformer-based models~\cite{vaswani2017, Karita2019} and provide step-by-step recipes for feature extraction, data preprocessing, model training, and inference for researchers to reproduce the benchmarks. Though there exist several counterparts, such as \textit{Lingvo}~\cite{Shen2019}, \textit{fairseq-ST}~\cite{Wang2020} and \textit{Kaldi}~\footnote{https://kaldi-asr.org/} style~\textit{ESPnet-ST}~\cite{inaguma2020},  \method is specially designed for speech translation tasks, which encapsulates the details of speech processing and frees the developers from data engineering. It is easy to use and extend.
The contributions of this work are as follows: 

\begin{compactitem}
    \item \method is designed specifically for end-to-end ST, with clean and simple code. It is lightweight and independent of \textit{Kaldi}, which simplifies installation and usage, and is more compatible for NLP researchers.
    \item We report strong benchmarks with well-designed hyper-parameters and show best practice on several ST corpora. We provide a series of recipes to reproduce them, which serves as reliable baselines for the speech translation field.  

\end{compactitem}



%% file: 02-design.tex
\section{Design and Features}
 \method is implemented with both TensorFlow2 and PyTorch backends. In this section, we will introduce the design components and features of this toolkit.

\subsection{Design}
\method divides one running job into four components: \texttt{Dataset}, \texttt{Model}, \texttt{Task} and \texttt{Executor}.

\paragraph{Dataset}
\method abstracts out a common interface \texttt{Dataset} for data input. For example, we can train a speech translation model from either a raw dataset tarball or pre-extracted record files. The \texttt{Dataset} iterates on the data files and standardizes the read records, e.g., ST tasks only accept key-value pairs storing audio signals/features and translations. One can implement their logic to accept the data of various modalities.

\paragraph{Model}
\method provides an optimal implementation of Transformer and its adaptation to speech-to-text tasks, which achieve state-of-the-art performance on standard benchmarks. Moreover, one can customize various models using TensorFlow2/PyTorch APIs or combine the encoders, decoders, and layers inside the \method.

\paragraph{Task}
\method abstracts out \texttt{Task} interface to bridge \texttt{Dataset} and \texttt{Model}. In detail, \texttt{Task} defines data pipelines to match the data samples from \texttt{Dataset} to the input formats of \texttt{Model}. For examples, ST task does tokenization on the text translations and transforms each token to index. In this way, user-defined \texttt{Dataset} and \texttt{Model} can be efficiently integrated into \method, as long as they share the same \texttt{Task}.

\paragraph{Executor}
\method provides the execution logic for handling basic workflows of training, validation, and inference. Researchers can either define their specific process of training and evaluation, or pay less attention to API details in \texttt{Executor} but reuse them by simply customizing \texttt{Dataset}, \texttt{Model} and \texttt{Task}. 

\subsection{Features}

\paragraph{Computation}
\method has high computation efficiency and it can be further optimized by enabling mixed-precision~\cite{Micikevicius2018} and XLA~(Accelerated Linear Algebra). Furthermore, \method supports fast distributed training using \textit{Horovod}~\cite{Sergeev2018} and \textit{Byteps}~\cite{Peng2019,Jiang2020} on large-scale scenarios.

\paragraph{Data Preprocessing}
\method supports on-the-fly data preprocessing via a number of lightweight python packages, like {python\_speech\_features}\footnote{https://github.com/jameslyons/python\_speech\_features} for extracting audio features (e.g. mel-frequency cepstral coefficients and log-mel filterbank coefficients). And for text processing, \method integrates some effective tokenizers, including moses tokenizer\footnote{The python version: https://github.com/alvations/sacremoses}, byte pair encoding (BPE)~\cite{sennrich2016} and SentencePiece\footnote{https://github.com/google/sentencepiece}.
Alternatively, the training data can be preprocessed and stored in binary files (e.g., TFRecord) beforehand, which is guaranteed to improve the I/O performance during training. Moreover, to simplify such operations, \method provides the command-line tool to create such record files, which automatically iterates on various data formats defined by \texttt{Dataset}, preprocesses data samples according to \texttt{Task} and writes to the disk. 

\paragraph{Transfer Learning}
\method supports initializing the model variables from well-trained models as long as they have the same variable names. As for ST, we can initialize the ST encoder with a well-trained ASR encoder and initialize the ST decoder with a well-trained MT decoder, which facilitates to achieve promising improvements. 
Besides, \method also provides scripts for converting released models from other repositories, like wav2vec2.0~\cite{Baevski2020w2v} and BERT~\cite{Devlin2019bert}. Researchers can conveniently integrate these pre-trained components to the customized models.

\paragraph{Simultaneous Translation} \method keeps up with the recent progress of simultaneous translation. The models are extended to train with streaming audio or text input.

\paragraph{Validation while Training}
\method supports customizing validation process during training. By default, \method offers evaluation on development data during training and keeps track of the checkpoints with the best evaluation results.

\paragraph{Monitoring}
\method supports TensorBoard for monitoring metrics during training, such as training loss, training speed, and evaluation results. 

\paragraph{Model Serving}
There is no gap between the research models and production models under \method, while they can be easily served with TensorFlow Serving. Moreover, for higher performance serving of standard transformer models, \method is able to integrate with other optimized inference libraries, like \textit{lightseq}~\cite{lightseq2020}.

%% file: 03-experiment.tex
\section{Speech Translation Benchmarks}

We conducted experiments on several benchmark speech translation corpora using \method and compared the performance with other open-source codebases and studies. Though that would be an unfair comparison due to the different model structures and hyperparameters, the goal of \method is to provide strong and reproducible benchmarks for future research.

\begin{table}[!t]\small
  \centering
  \begin{tabular}{ccccc}
    \toprule
    task & init scale & end scale & decay at & decay steps     \\
    \midrule
    MT & 1.0 & 1.0 & - & - \\
    ASR & 3.5 & 2.0 & 50k & 50k \\
    ST & 3.5 & 1.5 & 50k & 50k \\
    \bottomrule
  \end{tabular}
  \caption{Hyperparameters of the learning rate schedule. Take the case of ST, the learning rate is scaled up by 3.5x for the first 50k steps. Then, we linearly decrease the scaling factor to 1.5 for 50k steps.}
  \label{tb:lrschedule}
\end{table}

\subsection{Datasets}
We choose the following publicly available speech translation corpora that include speech in a source language aligned to text in a target language:

\noindent \textbf{\textit{libri-trans}}~\cite{Kocabiyikoglu2018} \footnote{https://github.com/alicank/Translation-Augmented-LibriSpeech-Corpus} is a small \textsc{EN$\rightarrow$FR} dataset which was originally started from the \textit{LibriSpeech} corpus, the audiobook recordings for ASR~\cite{Panayotov2015}. The English utterances were automatically aligned to the e-books in French, and 236 hours of English speech aligned to French translations at utterance level were finally extracted. It has been widely used in previous studies. As such, we use the clean 100-hour portion plus the augmented machine translation from Google Translate as the training data and follow its split of dev and test data.

\noindent \textbf{MuST-C}~\cite{gangi2019}\footnote{https://ict.fbk.eu/must-c/} is a multilingual speech translation corpus from English to 8 languages: Dutch (NL), French (FR), German (DE), Italian (IT), Portuguese (PT), Romanian (RO), Russian (RU) and Spanish (ES). MuST-C comprises at least 385 hours of audio recordings from English TED talks with their manual transcriptions and translations at sentence level for training, and we use the \textit{dev} and \textit{tst-COMMON} as our development and test data, respectively.
To the best of our knowledge, MuST-C is currently the largest speech translation corpus available for each language pair.

\input{03tbl-libritrans}

\input{03tbl-mustc}

\subsection{Data Preprocessing}
Beyond the officially released version, we performed no other audio to text alignment and data cleaning on \textit{libri-trans} and MuST-C datasets.

For speech features, we extracted 80-channel log-mel filterbank coefficients 
with windows of 25ms and steps of 10ms, resulting in 80-dimensional features per frame. The audio features of each sample were then normalized by the mean and the standard deviation.  
All texts were segmented into subword level by first applying Moses tokenizer and then BPE. 
In detail, we removed all punctuations and lowercased the sentences in the source side while the cases and punctuations of target sentences were reserved. The BPE rules were jointly learned with 8,000 merge operations and shared across ASR, MT, and ST tasks.

\footnotetext{\texttt{multi-bleu-detok.perl} in https://github.com/ \\ espnet/espnet/blob/master/utils/score\_bleu.sh}

\subsection{Benchmark Models}

We implemented Transformer~\cite{vaswani2017}, the state-of-the-art sequence-to-sequence model, for all our tasks.

In detail, for MT in cascade systems, the model included 6 layers for both encoder and decoders. The embedding dimension was 256, and the size of hidden units in feedforward layer was 2,048. The attention head for self-attention and cross-attention was set to 4. We used Adam optimizer~\cite{kingma2015} with $\beta_1=0.9,\beta_2=0.98$ and applied the same schedule algorithm as \citet{vaswani2017} for learning rate. 
We trained the MT models with a global batch size of 25,000 tokens.

As for ASR/ST, we referred to the recent progress of Transformer-based end-to-end ASR models~\cite{dong2018, Karita2019} and extended the basic transformer model to be compatible with audio inputs. The audio frames were first compressed by two-layer CNN with 256 channels, $3\times 3$ kernel and stride size 2, each of which was followed by a layer normalization. 
Then, we performed a linear transformation on the compressed audio representations to match the width of the transformer model. We used the same model structure as MT, except that we enlarged the number of encoder layers to 12 to obtain better performance. 
This configuration is labeled as \textit{transf-s} (transformer small).
For training, we used the same Adam optimizer as MT but set the warmup steps to 25,000, and we empirically scaled up the learning rate to accelerate the convergence. 
The hyperparameters of the learning rate schedule are listed in Table \ref{tb:lrschedule}.
Moreover, for GPU memory efficiency, we truncated the audio frames to 3,000 and removed training samples whose transcription length exceeded 120 and 150 for ASR and ST, respectively. 
The ASR models were trained with 120,000 frames per batch, while the batch size for ST was 80,000 frames. To further improve the performance of ST, we applied SpecAugment technique~\cite{Daniel2019} with frequency masking ($mF=2, F=27$) and time masking ($mT=2, T=70, p=0.2$).

Additionally, we applied label smoothing of value 0.1 for training all three tasks. The encoder of the ST model is initialized by the ASR encoder by default unless noted.

\subsection{Evaluation}
For evaluation, we averaged the latest 10 checkpoints and used a beam width of 4 with no length penalty for all the above tasks.

We use word error rate~(WER) to evaluate ASR models and report case-sensitive detokenized BLEU\footnote{https://github.com/mjpost/sacrebleu} for MT and ST models. In order to compare with existing works, we also report case-insensitive tokenized BLEU using \texttt{multi-bleu.perl} in Moses for \textit{libri-trans} dataset.

\subsection{Main Results}

\begin{table}[!t]\small
  \centering
  \begin{tabular}{lcc}\toprule
      Model & \ tok\   & detok  \\
      \midrule
      \textbf{Cascade} \\
      NeurST ASR \textit{transf-s} $\rightarrow$ MT & 17.4 & 16.0 \\ 
      \midrule
      \textbf{End-to-End\ } \\
      NeurST ST \textit{transf-s} & 17.8 & 16.3 \\ 
      ST \textit{transf-base} + AFS${^{t,f}}^\diamondsuit$ & 18.6 & 17.2 \\
    \bottomrule
    \end{tabular}
  \caption{Case-sensitive BLEU scores on \textit{libri-trans} test set under constrained setting. $\,^\diamondsuit$is from~\citet{Zhang2020} with the proposed adaptive feature selection method, which uses the transformer base setting (embedding size=512).}
  \label{tb:st_libri_sensitive}
\end{table}

The overall results and comparisons with other studies are illustrated in Table~\ref{tb:st_auglibri_lc} and \ref{tb:exp_mustc}. It is worth noting that all results are from single models rather than ensemble models.

To make a fair comparison on \textit{libri-trans} corpus, we list both tokenized and detokenized BLEU scores in Table~\ref{tb:st_auglibri_lc} and strive to distinguish the metric of existing literature.
Our transformer-based ST model, which only applies ASR pre-training and SpecAugment, achieves superior results versus recent works about knowledge distillation~\cite{liu2019end}, curriculum pre-training~\cite{Wang2020curr}, and LUT~\cite{dong2020ted}. 
Compared with the counterpart \textit{ESPnet-ST}, we also outperform by 0.5 BLEU, even though \citet{inaguma2020} apply additional techniques like speed perturbation, pre-trained MT decoder, and CTC loss for ASR pre-training. 
The cascade baseline is slightly worse than that of \textit{ESPnet-ST} (-0.2 BLEU) because the ASR+CTC can achieve lower WER (6.4)\footnote{from https://github.com/espnet/espnet/blob/master/egs/ \\ libri\_trans/asr1/RESULTS.md} while our pure end-to-end ASR obtains 8.8. We surprisingly find that the end-to-end ST model exceeds the cascade system by 0.4$\sim$0.5 BLEU. We will discuss this in detail in section \ref{sec:cas_vs_e2e}. And as a supplementary benchmark, we present case-sensitive BLEU scores in Table~\ref{tb:st_libri_sensitive}.

Table~\ref{tb:exp_mustc} illustrates the results on MuST-C \textit{tst-COMMON}. 
The results of our end-to-end ST model are competitive with both \textit{fairseq-ST} and \textit{ESPnet-ST}.

\input{03tbl-ablation.tex}

\subsection{Ablation Study}
Training a direct ST model is more complicated than training an ASR or MT model. Our preliminary experiment based on a pure end-to-end ST model fails to converge on \textit{libri-trans} corpus, which can be the result of the data scarcity. To alleviate this problem, pre-training some parts of the neural network is the most effective way and has been validated in all existing end-to-end ST studies. We show our results in Table~\ref{tb:st_libri_ablation} and \ref{tb:st_mustc_ablation} as a reference for future works. It turns out that we can obtain a reasonable or even better BLEU score by simply initializing the ST encoder with a pre-trained ASR encoder. The improvement by MT decoder initialization is relatively marginal in our setup. Furthermore, the SpecAugment technique can consistently boost ST models. 

\subsection{Cascade versus End-to-End}\label{sec:cas_vs_e2e}
Previous experiments on \textit{libri-trans} and MuST-C NL/PT show that the end-to-end systems have outperformed the cascade systems. Here we argue that the performance of the cascade systems above is hampered by a lack of quantitative data, and they should take advantage of large amounts of ASR and MT data separately.
Hence, we further extended \method to large-scale scenarios and experimented on the allowed datasets for IWSLT 2021 evaluation campaign\footnote{https://iwslt.org/2021/offline}. We followed the practice of~\citet{Zhao2021iwslt} to build our large cascade and end-to-end ST systems, which contains large-scale back-translation~\cite{sennrich2016bt} and pseudo labeling (also known as knowledge distillation) technologies. 
The results are illustrated in Table ~\ref{tb:st_cas_vs_e2e}. As seen, there is a significant loss of 1.7 BLEU between end-to-end ST and cascade ST. And the cascade system would have the potential to narrow the gap to the pure MT system by introducing extra punctuation restoration and true-case modules.

Though the cascade system is superior under large data conditions, we believe future researches on self-supervised learning, knowledge distillation, and dataset construction would realize the potential of end-to-end models.

\begin{table}[t]
    \centering
    \tabcolsep 2pt
    \begingroup
    \small
    \begin{tabular}{lc}\toprule
      Model & BLEU    \\
      \midrule
      large MT (w/ punc. \& cased) & 36.2  \\
      large MT (w/o punc.\& lc) & 34.3 \\
      large cascade ST & 31.4 \\
      large end-to-end ST & 29.7 \\
    \bottomrule
    \end{tabular}
    \caption{Case-sensitive detokenized BLEU scores on MuST-C EN-DE \textit{tst-COMMON}.}
    \label{tb:st_cas_vs_e2e}
    \endgroup
\end{table}

%% file: 03tbl-libritrans.tex
\begin{table*}[t]
    \centering
    \tabcolsep 2pt
    \begingroup
    \small
    \begin{tabular}{llcc}\toprule
      \multicolumn{2}{c}{Model} & \ tok \   & \  detok \   \\
      \midrule
      \multirow{2}{*}{Cascade\quad} 
      & \textit{ESPnet-ST} ASR \textit{transf-s} + CTC $\rightarrow$ MT~\cite{inaguma2020}$^\dagger$ & - & 17.0 \\
      \cmidrule{2-4}
      & \textbf{NeurST} ASR \textit{transf-s} $\rightarrow$ MT & 18.2 & 16.8 \\ 
      \midrule
        \multirow{9}{*}{End-to-End} 
            & ST BiLSTM~\cite{Bahar2019} & 17.0 & 16.2 \\
            & ST \textit{transf-s}~\cite{liu2019end}  &  14.3 & - \\
            & ST \textit{transf-s} + KD~\cite{liu2019end} & 17.0 & - \\
            & \textit{ESPnet-ST} ST \textit{transf-s}~\cite{inaguma2020}$^\dagger$ & - & 16.7 \\
            & TCEN-LSTM~\cite{wang2020bridging}$^\flat$ & - & 17.1 \\
            & ST \textit{transf-s}~\cite{Wang2020curr} & 16.0 & - \\
            & ST \textit{transf-s} + curriculum pre-training~\cite{Wang2020curr} & 17.7 & - \\
            & LUT~\cite{dong2020ted} & 17.8 & - \\
            \cmidrule{2-4}
            & \textbf{NeurST} ST \textit{transf-s} & 18.7 & 17.2 \\ 
    \bottomrule
    \end{tabular}
    \caption{Case-insensitive BLEU scores on \textit{libri-trans} test set under constrained setting (without additional ASR and MT data). $^\dagger$Notably, we refer to the results presented in \texttt{espnet/egs/libri\_trans/st1} and consider them as detokenized BLEU according to the evaluation script in the repository\footnotemark. $^\flat$ The result of TCEN-LSTM is also marked as detokenized BLEU due to its implementation on \textit{ESPnet-ST}.}
    \label{tb:st_auglibri_lc}
    \endgroup
\end{table*}

%% file: 03tbl-mustc.tex
\begin{table*}[t]
    \centering
    \tabcolsep 2pt
    \begingroup
    \small
    \begin{tabular}{llccccccccc}\toprule
      \multicolumn{2}{c}{Model} & DE & ES & FR & IT & NL & PT & RO & RU & avg. \\
      \midrule
      \multirow{2}{*}{Cascade} 
      & \tabincell{l}{ESPnet-ST ASR \textit{transf-s} + CTC $\rightarrow$ MT\\ \cite{inaguma2020}} & \ 23.7 \  & \ 28.7 \  & \  33.8 \  & \ 24.0 \ & \ 27.9\  & \ 29.0\  & \ 22.7 \  & \  16.4\ & \ 25.8 \  \\
      \cmidrule{2-11}
      & \textbf{NeurST} ASR \textit{transf-s} $\rightarrow$ MT & 23.4	& 28.0 & 33.9 & 23.8 & 27.1 & 28.3 & 22.2 & 16.0 & 25.3 \\ 
      \midrule
        \multirow{4}{*}{End-to-End} 
            & \textit{ESPnet-ST} ST \textit{transf-s}~\cite{inaguma2020} & 22.9 & 28.0 & 32.8 & 23.8 & 27.4 & 28.0 & 21.9 & 15.8 & 25.1 \\
            & \textit{fairseq-ST} ST \textit{transf-s}~\cite{Wang2020} & 22.7 & 27.2 & 32.9 & 22.7 & 27.3 & 28.1 & 21.9 & 15.3 & 24.8 \\
            & ST \textit{transf-base} + AFS$^{t,f}$~\cite{Zhang2020} & 22.4 & 26.9 & 31.6 & 23.0 & 24.9 & 26.3 & 21.0 & 14.7 & 23.9  \\
            \cmidrule{2-11}
            & \textbf{NeurST} ST \textit{transf-s} & 22.8 & 27.4 & 33.3  & 22.9 & 27.2 & 28.7 & 22.2 & 15.1 & 24.9 \\
    \bottomrule
    \end{tabular}
    \caption{Case-sensitive detokenized BLEU scores on MuST-C \textit{tst-COMMON}. } \label{tb:exp_mustc}
    \endgroup
\end{table*}

%% file: 03tbl-ablation.tex
\begin{table}[t]
    \centering
    \tabcolsep 2pt
    \begingroup
    \small
    \begin{tabular}{lcc}\toprule
      Model & \  NeurST \  & \  \textit{ESPnet-ST}\    \\
      \midrule
      ST + ASR enc init. & 16.5 & 15.5 \\
      \quad + MT dec init. & 16.6 & 16.2 \\
      \qquad + SpecAug. & 17.2 & 16.7 \\
      ST + ASR enc init. + SpecAug. & 17.2 & - \\
    \bottomrule
    \end{tabular}
    \caption{Case-insensitive detokenized BLEU scores on \textit{libri-trans} test set with difference setups.}
    \label{tb:st_libri_ablation}
    \endgroup
\end{table}

\begin{table}[t]
    \centering
    \tabcolsep 2pt
    \begingroup
    \small
    \begin{tabular}{lcc}\toprule
      Model & \  NeurST \  & \  \textit{ESPnet-ST}\    \\
      \midrule
      pure ST & 18.6 & - \\
      \, + ASR enc init. & 21.9 & 21.8 \\
      \, \,  + MT dec init. & 22.1 & 22.3 \\
       \, \, \, + SpecAug. & 23.3 & 22.9 \\
      ST + ASR enc init. + SpecAug. & 22.8 & - \\
    \bottomrule
    \end{tabular}
    \caption{Case-sensitive detokenized BLEU scores on MuST-C EN-DE \textit{tst-COMMON}  with difference setups.}
    \label{tb:st_mustc_ablation}
    \endgroup
\end{table}

%% file: 04-conclusion.tex
\section{Conclusion}
We introduce \method toolkit for easily building and training end-to-end speech translation models. We provide straightforward recipes for audio data pre-processing, training, and inference, which we believe is friendly with NLP researchers. Moreover, we report strong and reproducible benchmarks and will continuously catch up on advanced progress using \method, which can be regarded as the reliable baselines for the ST field.